\RequirePackage{cmap}
\documentclass{elsarticle}

\usepackage{graphicx}
\usepackage{multirow}
\usepackage{amsmath}
\usepackage{amssymb}
\usepackage{xcolor}
\usepackage{booktabs}
\usepackage{tabularx}
\usepackage{placeins}
\usepackage{hyperref}
\hypersetup{hidelinks}
\newcommand{\NRecipients}{2,485}
\newcommand{\NDonors}{1,249}
\newcommand{\NSamples}{25,624}

\newcommand{\NSceneTrain}{8}
\newcommand{\NSceneTest}{4}

\newcommand{\QCDropPct}{0.4\%}
\newcommand{\MetaAUROCvOne}{0.650}
\newcommand{\MetaAUROCvTwo}{0.612}
\newcommand{\SegPixFOne}{0.821}
\newcommand{\SegPixFOneAll}{0.621}
\newcommand{\ZSAUROCMin}{0.504}
\newcommand{\ZSAUROCMax}{0.745}
\newcommand{\ZSBest}{B-Free$^\ddagger$}
\newcommand{\ZSCtrlFPRMin}{0.55}
\newcommand{\ZSCtrlFPRMax}{1.00}
\newcommand{\ZSMCCMin}{-0.42}

\newcommand{\FTMesorchID}{0.946}

\newcommand{\FTBGRIFTID}{0.945}

\newcommand{\FTMVSSID}{0.917}
\newcommand{\FTSparseID}{0.838}
\newcommand{\BFDNetID}{0.797}
\newcommand{\SPDNetID}{0.669}
\newcommand{\FTMesorchCtrl}{0.060}
\newcommand{\FTBGRIFTCtrl}{0.099}

\newcommand{\BFDNetCtrl}{0.38}
\newcommand{\SPDNetCtrl}{0.95}
\newcommand{\NoPretrainID}{0.772}
\newcommand{\NoResID}{0.951}
\newcommand{\MesorchZSFOnePaired}{0.892}
\newcommand{\MesorchZSFThreePaired}{0.549}
\newcommand{\TruForZSFOnePaired}{0.914}
\newcommand{\TruForZSFThreePaired}{0.587}
\newcommand{\FTMesorchFFivePaired}{0.777}
\newcommand{\FTMesorchMCFFivePaired}{0.860}
\newcommand{\FTMesorchFFivePixF}{0.436}
\newcommand{\SegFFivePixF}{0.080}
\newcommand{\ZSYoudenTPRMin}{0.21}
\newcommand{\ZSYoudenTPRMax}{0.67}
\newcommand{\ZSYoudenFPRMin}{0.07}
\newcommand{\ZSYoudenFPRMax}{0.29}
\newcommand{\DropGenMin}{4.3}
\newcommand{\DropGenMax}{5.6}

\newcommand{\DropBGMax}{2.0}

\newcommand{\DropSubjMax}{2.1}

\newcommand{\DropRobustMax}{2.1}
\newcommand{\StrongGOnePairedMin}{0.81}
\newcommand{\StrongGOnePairedMax}{0.87}

\newcommand{\StrongGTwoPairedMin}{0.98}
\newcommand{\StrongGTwoPairedMax}{1.00}

\newcommand{\StrongGThreePairedMin}{0.82}
\newcommand{\StrongGThreePairedMax}{0.88}

\newcommand{\NZeroShotLower}{seven}
\newcommand{\BFreeID}{0.745}
\newcommand{\BFreeFPRAZero}{0.35}
\newcommand{\BFreeFPRCOne}{0.32}
\newcommand{\BFreeFPRCTwo}{0.94}
\newcommand{\BFreeFPRCThree}{0.44}

\newcommand{\BFreeFFivePaired}{0.113}

\newcommand{\BFreeGAuthMin}{0.92}
\newcommand{\BFreeGAuthMax}{1.00}
\newcommand{\BFreeGPairedMin}{0.77}
\newcommand{\BFreeGPairedMax}{0.88}

\newcommand{\MetaFlagAuthOne}{29.8\%}
\newcommand{\MetaFlagCtrlOne}{100.0\%}
\newcommand{\MetaFlagAuthTwo}{88.9\%}
\newcommand{\MetaFlagCtrlTwo}{89.8\%}
\newcommand{\MCBGRIFTSubjAUROC}{+0.023}
\newcommand{\MCBGRIFTSubjAUROCCI}{[+0.007, +0.040]}
\newcommand{\MCBGRIFTSubjAUROCPHolm}{0.004}
\newcommand{\MCBGRIFTSubjFPR}{-0.118}
\newcommand{\MCBGRIFTSubjFPRCI}{[-0.245, -0.012]}
\newcommand{\MCBGRIFTSubjFPRPHolm}{0.073}
\newcommand{\MCBGRIFTGenAUROC}{-0.009}
\newcommand{\MCBGRIFTGenAUROCCI}{[-0.032, +0.010]}

\newcommand{\MCMesorchGenAUROC}{-0.004}
\newcommand{\MCMesorchGenAUROCCI}{[-0.032, +0.037]}

\newcommand{\MCMesorchFFive}{+0.083}
\newcommand{\MCMesorchFFiveCI}{[+0.035, +0.135]}
\newcommand{\MCMesorchFFivePHolm}{0.004}


\journal{Journal of Information Security and Applications}

\begin{document}

\begin{frontmatter}

\title{Image Forgery Detection under Generative and Composited Background Replacement: A Benchmark with Processing-Matched Controls}

\author[1]{Bugra Alperen Uluirmak\corref{cor1}\fnref{orcid1}}
\ead{uluirmak@erciyes.edu.tr}
\author[2]{Rifat Kurban\fnref{orcid2}}
\ead{rifat.kurban@agu.edu.tr}

\cortext[cor1]{Corresponding author}
\fntext[orcid1]{ORCID: 0009-0000-3077-673X}
\fntext[orcid2]{ORCID: 0000-0002-0277-2210}

\affiliation[1]{organization={Erciyes University},
            city={Kayseri},
            country={T\"{u}rkiye}}
\affiliation[2]{organization={Abdullah G\"{u}l University},
            city={Kayseri},
            country={T\"{u}rkiye}}

\begin{abstract}
Consumer tools can now keep the subject of a photograph and replace, relight or edit everything behind it. Such edits place genuine people and objects in scenes they were never in, yet involve processing---resampling, harmonization, diffusion autoencoders---that also occurs in benign workflows. We introduce BG-REAL, a benchmark of \NSamples{} images built from \NRecipients{} Open Images photographs with a dominant subject and real donor scenes. Each recipient is edited by donor compositing, learned harmonization, generative background inpainting, foreground-conditioned relighting and local background edits, and is paired with authentic controls that share the processing but not the semantic edit; all images of a recipient share file format and JPEG quality. Three generators (including FLUX.1 Fill), scene categories and subject classes are held out. Seven detectors published between 2022 and 2026 reach zero-shot AUROCs of at most \ZSAUROCMax{} and have no useful operating point: manipulation localisers mark the preserved subject instead of the replaced background, and a detector of AI-generated images flags most authentic images passed through a diffusion autoencoder. After fine-tuning, the strongest detectors reach an AUROC of about 0.95, while re-implemented background-forgery detectors remain below 0.80. Controls show that full-background replacements by the training generators are detected from the edit itself, whereas local edits and SDXL and FLUX edits remain hard, and that full-background localisation largely reduces to subject segmentation. A model-agnostic training objective that uses the controls improves detection on unseen subjects for one detector and on local edits for another without changing in-distribution accuracy. The data are released.

\end{abstract}


\end{frontmatter}

\section{Introduction}
\label{sec:intro}

Replacing the background of a photograph while keeping its subject is now a one-click operation: video-conferencing software swaps backgrounds in real time, phone galleries offer ``magic eraser'' and AI background replacement, and relighting tools re-render the subject so that it matches a new scene. The resulting images keep a genuine person, vehicle or object but place it in a scene it was never in. For forensic analysis this setting differs from classical splicing in two ways. First, the manipulated area is usually the larger part of the image, so detectors that look for a small foreign object have little to latch onto. Second, the editing pipelines involve processing steps---resampling, alpha blending, learned colour harmonization, passing the whole image through the autoencoder of a diffusion model---that also occur in benign workflows. A detector can therefore appear accurate by reacting to the processing rather than to the fact that the background was replaced.

\paragraph{Threat model} We consider an adversary who keeps a genuine subject and changes where it appears to be, in order to fabricate an alibi, misattribute a location or event, or support disinformation; the adversary uses widely available tools and may re-encode the result. The analyst receives a single image without provenance metadata and must decide whether its background was manipulated, ideally indicating where. Two errors matter in practice. Missing a manipulation lets a fabricated context pass; flagging a benign image---one that was merely resampled, recompressed, colour-corrected or passed through an editing app without changing its content---undermines the credibility of the analysis and can wrongly discredit genuine evidence. A detector is only useful in this setting if it separates the semantic edit from the processing that accompanies it.

Existing resources cover parts of this problem. General image-manipulation benchmarks \cite{chen2013casia,wen2016coverage,novozamsky2020imd2020,mahfoudi2019defacto,ma2024imdlbenco} are dominated by object-level splicing and copy-move. Recent localized-generation datasets add diffusion-based inpainting of objects \cite{mareen2024tgif,mareen2026tgif2} and, in BR-Gen \cite{cai2026brgen}, of ``stuff'' and background regions. Background forgery has been studied as a detection problem with person-centric data \cite{elmaci2024bfdnet,elmaci2025spdnet}. None of these, however, controls for the processing that a background edit implies, and none holds out generators, scene types and subject classes jointly.

We introduce BG-REAL, a benchmark for subject-preserving background manipulation. Its design rests on three decisions.
(i)~\emph{Realistic editing techniques, not one generator.} Every recipient photograph is edited with five technique classes: donor-background compositing, learned harmonization \cite{guerreiro2023pctnet}, generative background inpainting \cite{rombach2022ldm}, foreground-conditioned relighting with a generated background \cite{zhang2025iclight}, and local background edits by object removal \cite{suvorov2022lama} or text-guided insertion.
(ii)~\emph{Processing-matched authentic controls.} For every processing path we add an authentic counterpart that passes through the same operations without a semantic edit---a resampled self-composite, an autoencoder round trip of each generator, a harmonizer pass---and all samples of one recipient share the same file format and JPEG quality. This turns the question ``does the detector see the edit or the processing?'' into a paired comparison.
(iii)~\emph{Held-out conditions by construction.} Splits are assigned per recipient before generation; scene categories, three generators with different autoencoders \cite{podell2024sdxl,razzhigaev2023kandinsky,bfl2024flux} and four animal subject classes appear only in dedicated test splits, and donor photographs are never shared between training and evaluation.

Our contributions are:
\begin{itemize}
  \item BG-REAL, \NSamples{} images built from \NRecipients{} Open Images V7 photographs \cite{kuznetsova2020openimages} with a dominant subject and \NDonors{} real scene photographs, with subject, semantic-edit and changed-pixel masks, full generation metadata and released splits (Section~\ref{sec:benchmark}).
  \item An evaluation protocol built around matched controls: control-paired AUROC, control false-positive rates, localisation that is also scored on authentic images, and diagnostics that expose file-level and segmentation shortcuts (Section~\ref{sec:protocol}).
  \item A comparison of detectors published between 2022 and 2026---TruFor \cite{guillaro2023trufor}, MVSS-Net \cite{dong2022mvss}, HiFi-Net \cite{guo2023hifi}, IML-ViT \cite{ma2023imlvit}, Mesorch \cite{zhu2025mesorch} and SparseViT \cite{su2025sparsevit}---zero-shot and after fine-tuning, the AI-generated-image detector B-Free \cite{guillaro2025bfree} zero-shot, together with re-implementations of the background- and subject-forgery detectors BFDNet \cite{elmaci2024bfdnet} and SPDNet \cite{elmaci2025spdnet} (Section~\ref{sec:results}).
  \item A model-agnostic matched-control training objective that uses the controls during training, and an analysis of its effect on BG-RIFT, a compact detector we propose as a reference, and on Mesorch (Section~\ref{sec:method}).
\end{itemize}

\section{Related Work}
\label{sec:related}

\paragraph{Image manipulation detection and localisation}
Detectors for image manipulation combine semantic RGB features with low-level traces. MVSS-Net supervises edge and noise views at multiple scales \cite{dong2022mvss}; PSCC-Net progressively refines spatio-channel correlations \cite{liu2022pscc}; CAT-Net learns JPEG compression artefacts \cite{kwon2021catnet}; ObjectFormer models object-level consistency with RGB and frequency features \cite{wang2022objectformer}; HiFi-Net organises forgery attributes hierarchically \cite{guo2023hifi}; TruFor fuses RGB with a learned noise fingerprint and outputs a reliability map \cite{guillaro2023trufor}. Transformer-based localisers followed: IML-ViT uses a plain vision transformer at high resolution \cite{ma2023imlvit}, Mesorch combines CNN and transformer branches at mesoscopic scale \cite{zhu2025mesorch}, SparseViT learns non-semantic features with sparse self-attention instead of hand-crafted filters \cite{su2025sparsevit}, and collaborative transformers with forensic attention were proposed recently \cite{zhang2026cotransformers}. IMDL-BenCo standardises training and evaluation of many of these models \cite{ma2024imdlbenco}; we use its implementations and released checkpoints where available. These detectors are trained and evaluated mostly on object-level splicing, copy-move and removal. Detectors aimed specifically at AI-generated local edits have appeared more recently, among them localisers built on multimodal large language models that also explain their decisions \cite{xu2025fakeshield,huang2025sida} and the forgery-amplification detector proposed with BR-Gen \cite{cai2026brgen}; we do not evaluate these (Section~\ref{sec:limitations}). Detectors of fully generated images are related: B-Free trains on content-preserving self-conditioned reconstructions and local inpaintings so that the detector does not rely on semantic shortcuts \cite{guillaro2025bfree}. We evaluate it zero-shot as an image-level reference.

\paragraph{Datasets for manipulated and locally generated images}
CASIA \cite{chen2013casia}, COVERAGE \cite{wen2016coverage}, IMD2020 \cite{novozamsky2020imd2020} and DEFACTO \cite{mahfoudi2019defacto} established pixel-level evaluation of classical manipulations, and challenging-manipulation benchmarks followed \cite{zhang2024cimd}. With diffusion models, datasets shifted to localized generation: TGIF and TGIF2 study text-guided inpainting forgeries and the effect of regenerating the full image through the generator \cite{mareen2024tgif,mareen2026tgif2}; OpenSDI targets diffusion-generated content in the open world \cite{wang2025opensdi}. Closest to our setting, BR-Gen \cite{cai2026brgen} contains 150,000 locally generated images in which ``stuff'' regions or the background---obtained by inverting object masks---are inpainted with GAN and diffusion inpainters, and is split randomly into training, validation and test sets. BG-REAL differs from BR-Gen in three respects. It includes manipulation techniques that do not regenerate pixels from scratch (compositing with a real donor photograph, learned harmonization) and one that also changes the subject (relighting); it pairs every processing path with an authentic control that undergoes the same processing; and its test splits hold out generators, scene categories and subject classes instead of sampling them from the training distribution. Section~\ref{sec:results} also shows that, when the whole background is replaced, localising the edit reduces to segmenting the subject, which affects how localisation scores on such data should be read.

\paragraph{Background and subject forgery}
BFDNet \cite{elmaci2024bfdnet} segments people with Mask R-CNN and decides from DenseNet-121 features of the foreground and background whether the background of a portrait was replaced; SPDNet \cite{elmaci2025spdnet} extends the idea to individual spliced persons with four EfficientNet-B4 streams over RGB and SRM residuals \cite{fridrich2012srm} and introduces the SPDE dataset. Both show that contrasting subject and background evidence is a useful cue. Both are image- or instance-level classifiers evaluated on person-centric data with compositing-based forgeries. BG-REAL extends this line to non-person subjects, to generative and relighting-based edits, to pixel-level localisation of local edits, and to matched controls; we re-implement both detectors and evaluate them under the same protocol as the other methods (Section~\ref{sec:methods-compared}).

\paragraph{Image composition, harmonization and relighting}
Making composites look natural is a long-standing problem \cite{xue2012realism,zhu2015realism,niu2021compositionsurvey}. Learned harmonization networks such as DoveNet \cite{cong2020dovenet} and PCT-Net \cite{guerreiro2023pctnet} adjust the colour of a pasted foreground to its new background, and IC-Light relights a foreground consistently with a generated background \cite{zhang2025iclight}. Generative backgrounds are produced with latent diffusion inpainting \cite{rombach2022ldm,podell2024sdxl,razzhigaev2023kandinsky}, and objects are removed with large-mask inpainters \cite{suvorov2022lama}. BG-REAL uses these tools as manipulation generators; as they improve, the colour and lighting inconsistencies exploited by earlier detectors weaken, which is why we include each of them as a separate family.

\section{The BG-REAL Benchmark}
\label{sec:benchmark}

\subsection{Task}
Given an image, a detector outputs an image-level score that the background has been manipulated and, optionally, a pixel map of the manipulated region. A manipulation is \emph{subject-preserving}: the main subject keeps its identity and geometry, while the background is replaced entirely or edited locally. Each sample carries three masks: the \emph{subject mask}, the \emph{semantic edit mask} (the region whose content was changed on purpose; the localisation target) and the \emph{changed-pixel mask} (every pixel whose value changed by more than 10 grey levels, which for relighting and harmonization also covers the subject).

\subsection{Source images}
\label{sec:sources}
Recipients are Open Images V7 training images \cite{kuznetsova2020openimages} whose largest instance mask is a dominant subject: a whitelisted class (people, vehicles, everyday objects, animals), predicted mask IoU of at least 0.8, a bounding box at least twice as large as any other annotated instance, a mask covering 10--60\% of the image as a single connected component, and limited contact with the image border. Animal classes (dog, cat, horse, bird) form the subject-OOD pool and never appear in training. Donors are Open Images photographs without a dominant annotated instance, classified with CLIP \cite{radford2021clip} into \NSceneTrain{} training and \NSceneTest{} held-out scene categories; a donor is accepted only if a scene category outscores a set of non-scene descriptions (close-ups, portraits, crowds, food, vehicles, text) with probability at least 0.5. All images are resized to a common canonical size (long side 768 pixels, both sides multiples of 8) so that resolution cannot separate families.

\subsection{Edit families and matched controls}
\label{sec:families}
Table~\ref{tab:families} lists the families. Full-background manipulations restore the original subject pixels with a feathered mask after editing, except relighting, where changing the subject's illumination is the point of the edit. Diffusion edits use scene prompts derived from the recipient's assigned scene category; local edits either remove an annotated background object with LaMa or insert an object into a background box with text-guided inpainting at native resolution. Each manipulated family has a matched authentic control that reproduces its processing without changing scene content: \textbf{C1} composites the subject onto a Lanczos-resampled copy of its own image (the counterpart of pasting onto a resampled donor), \textbf{C2} passes the image through the generator's autoencoder and restores the subject (the counterpart of every diffusion edit, with one variant per held-out generator), and \textbf{C3} applies the harmonizer to the authentic image.

\begin{table}[t]
\centering
\caption{Families in BG-REAL. ``Edit mask'' is the localisation target. Controls are authentic (label 0) and reproduce the processing of the families listed in the last column without a semantic edit.}
\label{tab:families}
\small
\begin{tabularx}{\linewidth}{@{}l>{\raggedright\arraybackslash}X l l@{}}
\toprule
Code & Operation & Edit mask & Control / controls \\
\midrule
A0 & Authentic recipient & empty & -- \\
C1 & Subject composited onto a resampled copy of its own image & empty & F1 \\
C2 & Autoencoder round trip (SD1.5; SDXL, Kandinsky and FLUX for G1--G3), subject restored & empty & F3, F4, F5, G1--G3 \\
C3 & PCT-Net harmonizer applied to the authentic image & empty & F2 \\
\midrule
F1 & Subject pasted onto a real donor photograph & background & C1 \\
F2 & F1 followed by learned harmonization (PCT-Net \cite{guerreiro2023pctnet}) & background & C3 \\
F3 & Background regenerated by SD1.5 inpainting \cite{rombach2022ldm}, subject restored & background & C2 \\
F4 & Subject relit and background generated by IC-Light \cite{zhang2025iclight} & background & C2 \\
F5 & Background object removed (LaMa \cite{suvorov2022lama}) or inserted (SD1.5) & local region & C2 \\
G1 & As F3 with SDXL inpainting \cite{podell2024sdxl} (test only) & background & C2 (SDXL) \\
G2 & As F3 with Kandinsky 2.2 \cite{razzhigaev2023kandinsky} (test only) & background & C2 (Kandinsky) \\
G3 & As F3 with FLUX.1 Fill [dev] \cite{bfl2024flux}, 4-bit weights (test only) & background & C2 (FLUX) \\
\bottomrule
\end{tabularx}
\end{table}

\begin{figure*}[t]
\centering
\includegraphics[width=\linewidth]{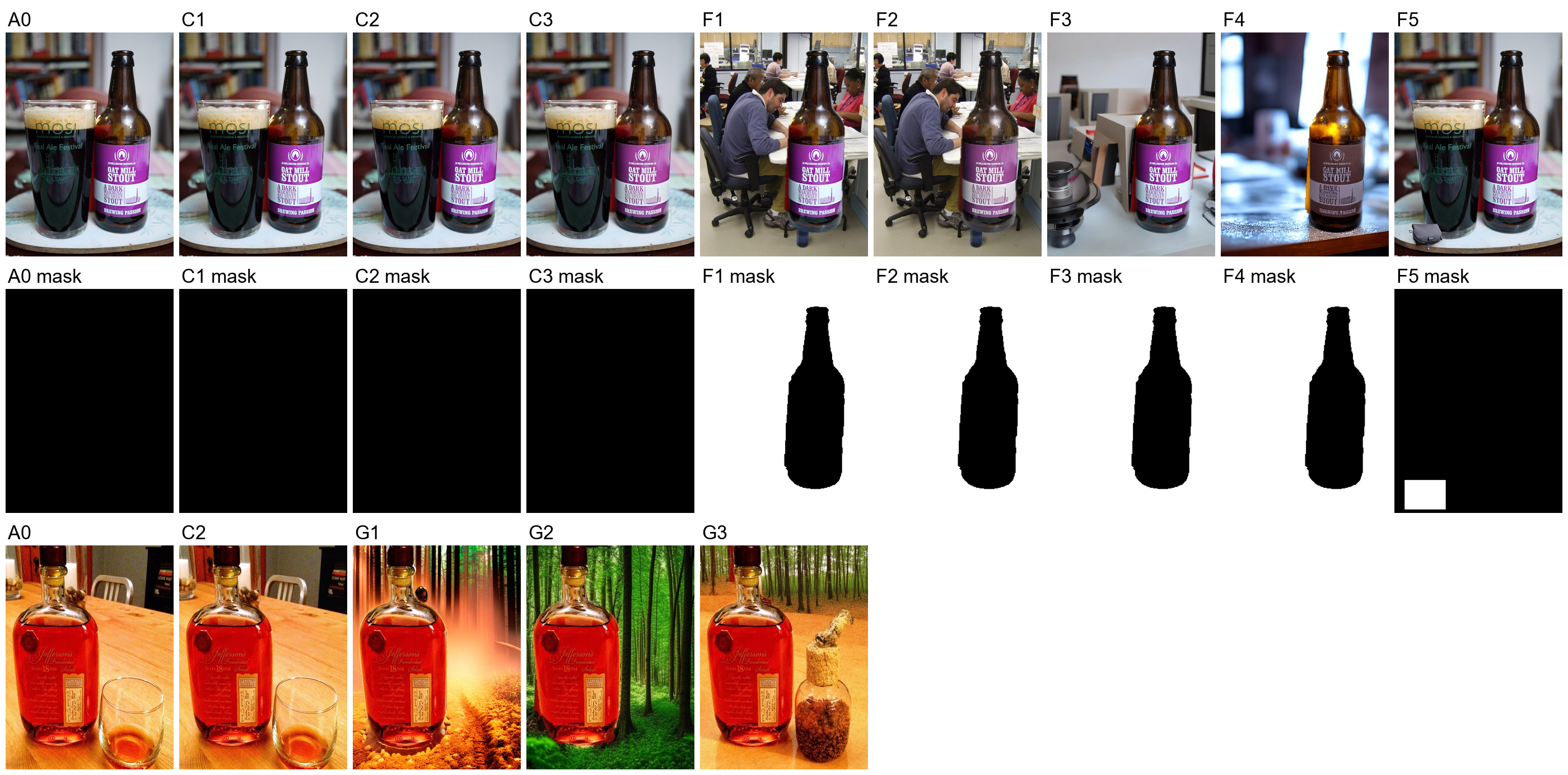}
\caption{One test-ID recipient in every family (top), the semantic edit masks used as localisation target (middle; white = edited), and a generator-OOD recipient with its SDXL (G1), Kandinsky (G2) and FLUX.1 Fill (G3) edits (bottom). Controls C1--C3 are visually indistinguishable from the authentic image. All manipulated images and controls were produced with the generative and harmonization models listed in Table~\ref{tab:families} as part of the benchmark; source photographs are from Open Images V7.}
\label{fig:families}
\end{figure*}

Figure~\ref{fig:families} shows one recipient across all families.

\subsection{Encoding, robustness and quality control}
After editing, every sample is written with a shared encoding policy: PNG with probability 0.2, otherwise JPEG with quality drawn uniformly from 75--95. The draw depends only on the recipient, so the authentic image, its controls and its manipulations are stored in exactly the same format and quality. A robustness variant (downscaling by a random factor in 0.5--0.9, one intermediate JPEG compression at quality 60--85, resizing back, optional sharpening) is applied identically to all families of every test-ID recipient and of a fixed 20\% of training recipients. Quality control is automatic: we discard black or flat outputs (including generations blanked by the safety checker), outputs whose restored subject interior deviates from the original (PSNR below 35~dB after eroding the subject mask by three pixels), and manipulations that change less than 30\% of their edit mask. In total \QCDropPct{} of generated samples were removed.

\subsection{Splits}
\label{sec:splits}
Splits are assigned per recipient before any generation, stratified by subject group, so that every family of a recipient lives in one split. Table~\ref{tab:splits} summarises them. Training, validation and test-ID recipients use training scene categories and generators. Background-OOD recipients use only held-out scene categories, both for donor photographs and for generation prompts. Generator-OOD recipients are edited only by SDXL, Kandinsky~2.2 and FLUX.1 Fill, whose autoencoders differ from the SD1.5 autoencoder used in training; FLUX.1 Fill is a rectified-flow transformer with a 16-channel autoencoder, released after the other generators, and runs on the 8~GB GPU with its transformer and T5 encoder quantised to 4 bits. Subject-OOD recipients have animal subjects. Donor photographs of the training categories are partitioned once into a training pool and an evaluation pool, so no donor appears both in training and in any evaluation split. The test-matched-control split collects the controls of test-ID recipients, and the robustness split contains the robustness variants of test-ID recipients.

\begin{table}[t]
\centering
\caption{Splits of BG-REAL (sample counts after quality control). The robustness and matched-control splits reuse the test-ID recipients, so recipient counts do not add up across rows.}
\label{tab:splits}
\small
\begin{tabular}{@{}lrrl@{}}
\toprule
Split & Recipients & Samples & Held-out factor \\
\midrule
train & 1,192 & 12,636 & -- \\
val & 200 & 1,795 & -- \\
test-id & 199 & 1,784 & -- \\
test-background-ood & 199 & 1,782 & scene categories \\
test-generator-ood & 198 & 1,380 & generators \\
test-subject-ood & 497 & 4,463 & subject classes \\
test-robustness & 199 & 1,784 & post-processing \\
test-matched-control & 199 & 597 & (controls of ID) \\
 
\bottomrule
\end{tabular}
\end{table}

\section{Evaluation Protocol}
\label{sec:protocol}

\paragraph{Detection} We report AUROC and true- and false-positive rates. The decision threshold of every method is chosen once by maximising F1 on the validation split and then fixed for all test splits. Because 55\% of validation images are manipulated, this criterion degenerates to flagging almost every image when a detector separates the classes poorly; we therefore also report a second operating point, the threshold that maximises the true-positive minus the false-positive rate (Youden's $J$) on the validation split. The released evaluation tables additionally contain AUPRC, balanced accuracy, expected calibration error \cite{nixon2019calibration} and 95\% confidence intervals from 1,000 bootstrap resamples of test images for every method and split.

\paragraph{Matched-control measures} For each manipulated family we compute the \emph{control-paired AUROC}: the AUROC between the manipulated images and their own matched controls on the same recipients. A detector that only reacts to processing scores near 0.5 here even if its ordinary AUROC (manipulated versus authentic) is high. We also report the false-positive rate of each control type at the fixed threshold.

\paragraph{Localisation} When the whole background is replaced, the semantic edit mask is the complement of the subject. A detector that merely segments the subject would then reach a perfect per-image pixel F1 on manipulated images (Section~\ref{sec:diagnostics}). We therefore report, besides per-image pixel F1, IoU, Matthews correlation and tolerant boundary F1 \cite{perazzi2016davis} on manipulated images, a \emph{pooled pixel F1} accumulated over all images of a split, in which pixels flagged on authentic images and controls count as false positives, and we analyse localisation separately on local edits (F5), where the edited region is not determined by the subject.

\paragraph{Diagnostics} \label{sec:diagnostics} Two non-forensic baselines bound what the data allows without forensic evidence. A \emph{metadata} baseline is a logistic regression on file-level properties (bits per pixel, format, mean JPEG quantisation step, width, height) trained on the training split. A \emph{segmentation} baseline predicts the complement of the ground-truth subject mask with a constant score.

\section{Methods}
\label{sec:method}

\subsection{Compared detectors}
\label{sec:methods-compared}
We evaluate official checkpoints zero-shot: TruFor \cite{guillaro2023trufor} and HiFi-Net \cite{guo2023hifi} with their released code, and MVSS-Net \cite{dong2022mvss}, IML-ViT \cite{ma2023imlvit}, Mesorch \cite{zhu2025mesorch} and SparseViT \cite{su2025sparsevit} with the IMDL-BenCo implementations \cite{ma2024imdlbenco} and the authors' released weights. B-Free \cite{guillaro2025bfree}, a detector of AI-generated images, is run with its released code and weights on five $504\times504$ crops; it gives only an image-level score. Each method uses its own input protocol (native resolution for TruFor and HiFi-Net, zero-padding to 1024 pixels for IML-ViT, resizing to 512 pixels otherwise). TruFor and HiFi-Net provide their own detection scores and MVSS-Net takes the maximum of its mask, as in the original; for Mesorch, SparseViT and IML-ViT, which have no image-level output, the image score is the mean of the top 1\% of pixel probabilities. MVSS-Net, Mesorch and SparseViT are additionally fine-tuned on the BG-REAL training split, starting from their released weights and keeping their original losses. BFDNet \cite{elmaci2024bfdnet} and SPDNet \cite{elmaci2025spdnet} are re-implemented from their papers (two DenseNet-121 streams, and four EfficientNet-B4 streams over RGB and SRM residuals with squeeze-and-excitation fusion) and trained on BG-REAL; they receive the ground-truth subject mask in place of their instance segmenter, which favours them. CAT-Net \cite{kwon2021catnet} and PSCC-Net \cite{liu2022pscc} could not be run in our environment: CAT-Net's DCT stream requires a compiled JPEG library, and PSCC-Net's weights are distributed only through a service we could not access.

\subsection{BG-RIFT}
BG-RIFT is a compact reference detector built around the structure of the task. An ImageNet-pretrained ConvNeXt-T \cite{liu2022convnext} encodes the RGB image; a ResNet-18 encodes fixed SRM residuals \cite{fridrich2012srm}; the two are fused at four scales. A feature pyramid feeds a three-way region head (subject, background, subject boundary), supervised with the subject mask, whose soft output conditions the mask head that predicts the semantic edit mask. The image head combines pooled deep features with statistics of the predicted mask. The loss is binary cross-entropy plus Dice for the mask, binary cross-entropy for the image label and cross-entropy with weight 0.5 for the region head.

\subsection{Matched-control training}
\label{sec:mc-training}
All trained detectors use the same batch construction: each batch consists of triplets $(x^{m}, x^{c}, x^{a})$ of a manipulated image, its matched control and the authentic image of the same recipient. On top of each detector's own loss we optionally add two terms on the image logits $z$ (for a detector without an image-level output, the logit of its top-1\% pooled mask score, so that the terms reach the mask head):
\begin{align}
\mathcal{L}_{\text{rank}} &= \max\bigl(0,\; \delta - (z(x^{m}) - z(x^{c}))\bigr), \\
\mathcal{L}_{\text{inv}} &= \bigl|\sigma(z(x^{c})) - \sigma(z(x^{a}))\bigr|,
\end{align}
with margin $\delta=1$. The ranking term asks the manipulated image to score above the image that underwent the same processing without the edit; the invariance term asks the processed control to score like the untouched original. Both terms use only image-level outputs and can be added to any detector. Because the batch construction is identical with and without these terms, their effect is isolated.

\subsection{Training details}
Images are resized to $512\times512$. We train with AdamW \cite{loshchilov2019adamw} (weight decay $10^{-4}$), a one-cycle learning-rate schedule, gradient clipping at norm 5, mixed precision (float16; bfloat16 for SPDNet, whose four EfficientNet streams overflow in float16) and horizontal flips. Detectors initialised from released forensic checkpoints (MVSS-Net, Mesorch, SparseViT) are fine-tuned for three epochs at a peak learning rate of $5\times10^{-5}$; detectors initialised from ImageNet weights are trained at $10^{-4}$ (BFDNet, SPDNet) or $2\times10^{-4}$ (BG-RIFT) for four epochs, except SPDNet, which we trained for three because of its cost. We select the checkpoint with the best validation AUROC and repeat every configuration with three seeds; all runs fit on one 8~GB GPU, with batch sizes chosen per detector and gradient accumulation to keep six images (two triplets) per optimisation step.

\section{Results}
\label{sec:results}

Table~\ref{tab:main} reports detection on every test split, Table~\ref{tab:paired} the control-paired AUROC of each family, Table~\ref{tab:controls} the false-positive rate of each control type, and Table~\ref{tab:loc} localisation. Fine-tuned results are means and standard deviations over three seeds; zero-shot results are deterministic.

\begin{table*}[t]
\centering
\caption{Detection AUROC on the test splits and false-positive rates (FPR) at the threshold fixed on the validation split: on the matched controls of test-ID recipients and on the authentic and control images of subject-OOD recipients. ZS: official checkpoints without BG-REAL training; fine-tuned: trained on the BG-REAL training split (mean $\pm$ std over three seeds). MC: matched-control terms. $^\dagger$Re-implementation, given the ground-truth subject mask. $^\ddagger$Detector of AI-generated images (image-level score only).}
\label{tab:main}
\small
\setlength{\tabcolsep}{4pt}
\resizebox{\linewidth}{!}{%
\begin{tabular}{@{}lccccccc@{}}
\toprule
 & \multicolumn{5}{c}{AUROC $\uparrow$} & \multicolumn{2}{c}{FPR $\downarrow$} \\
\cmidrule(lr){2-6}\cmidrule(lr){7-8}
Method & ID & BG-OOD & Gen-OOD & Subj-OOD & Robust & Controls & Subj-OOD \\
\midrule
\multicolumn{8}{@{}l}{\textit{zero-shot}} \\
TruFor & 0.717 & 0.749 & 0.645 & 0.717 & 0.609 & 0.824 & 0.856 \\
HiFi-Net & 0.504 & 0.533 & 0.518 & 0.465 & 0.502 & 0.992 & 0.995 \\
MVSS-Net & 0.613 & 0.635 & 0.587 & 0.619 & 0.587 & 0.995 & 0.971 \\
IML-ViT & 0.645 & 0.656 & 0.650 & 0.619 & 0.592 & 1.000 & 0.998 \\
Mesorch & 0.742 & 0.750 & 0.623 & 0.757 & 0.716 & 0.554 & 0.519 \\
SparseViT & 0.688 & 0.689 & 0.618 & 0.671 & 0.670 & 0.997 & 0.998 \\
B-Free$^\ddagger$ & 0.745 & 0.745 & 0.835 & 0.748 & 0.752 & 0.564 & 0.434 \\
\midrule
\multicolumn{8}{@{}l}{\textit{fine-tuned}} \\
MVSS-Net & 0.917\,{\scriptsize$\pm$0.010} & 0.914\,{\scriptsize$\pm$0.007} & 0.870\,{\scriptsize$\pm$0.004} & 0.914\,{\scriptsize$\pm$0.009} & 0.896\,{\scriptsize$\pm$0.001} & 0.125\,{\scriptsize$\pm$0.019} & 0.247\,{\scriptsize$\pm$0.022} \\
SparseViT & 0.838\,{\scriptsize$\pm$0.009} & 0.833\,{\scriptsize$\pm$0.009} & 0.816\,{\scriptsize$\pm$0.007} & 0.763\,{\scriptsize$\pm$0.026} & 0.782\,{\scriptsize$\pm$0.010} & 0.308\,{\scriptsize$\pm$0.012} & 0.508\,{\scriptsize$\pm$0.036} \\
Mesorch & 0.946\,{\scriptsize$\pm$0.001} & 0.939\,{\scriptsize$\pm$0.008} & 0.903\,{\scriptsize$\pm$0.030} & 0.941\,{\scriptsize$\pm$0.016} & 0.933\,{\scriptsize$\pm$0.007} & 0.060\,{\scriptsize$\pm$0.024} & 0.128\,{\scriptsize$\pm$0.061} \\
Mesorch + MC & 0.952\,{\scriptsize$\pm$0.005} & 0.933\,{\scriptsize$\pm$0.003} & 0.898\,{\scriptsize$\pm$0.006} & 0.954\,{\scriptsize$\pm$0.005} & 0.939\,{\scriptsize$\pm$0.005} & 0.059\,{\scriptsize$\pm$0.008} & 0.129\,{\scriptsize$\pm$0.011} \\
BFDNet$^\dagger$ & 0.797\,{\scriptsize$\pm$0.007} & 0.790\,{\scriptsize$\pm$0.010} & 0.799\,{\scriptsize$\pm$0.024} & 0.741\,{\scriptsize$\pm$0.014} & 0.796\,{\scriptsize$\pm$0.006} & 0.377\,{\scriptsize$\pm$0.032} & 0.619\,{\scriptsize$\pm$0.017} \\
SPDNet$^\dagger$ & 0.669\,{\scriptsize$\pm$0.025} & 0.677\,{\scriptsize$\pm$0.036} & 0.696\,{\scriptsize$\pm$0.020} & 0.660\,{\scriptsize$\pm$0.024} & 0.697\,{\scriptsize$\pm$0.024} & 0.946\,{\scriptsize$\pm$0.073} & 0.929\,{\scriptsize$\pm$0.090} \\
BG-RIFT & 0.945\,{\scriptsize$\pm$0.001} & 0.927\,{\scriptsize$\pm$0.007} & 0.902\,{\scriptsize$\pm$0.002} & 0.924\,{\scriptsize$\pm$0.012} & 0.935\,{\scriptsize$\pm$0.004} & 0.099\,{\scriptsize$\pm$0.012} & 0.353\,{\scriptsize$\pm$0.066} \\
BG-RIFT + MC & 0.949\,{\scriptsize$\pm$0.002} & 0.932\,{\scriptsize$\pm$0.002} & 0.893\,{\scriptsize$\pm$0.011} & 0.947\,{\scriptsize$\pm$0.004} & 0.938\,{\scriptsize$\pm$0.003} & 0.082\,{\scriptsize$\pm$0.017} & 0.236\,{\scriptsize$\pm$0.039} \\
 
\bottomrule
\end{tabular}}
\end{table*}

\subsection{Detectors trained on other data do not transfer}
\label{sec:res-zeroshot}
Without BG-REAL training, the \NZeroShotLower{} detectors reach test-ID AUROCs between \ZSAUROCMin{} and \ZSAUROCMax{} (\ZSBest{} is best), and the scores of the six manipulation localisers barely change on the held-out splits (Table~\ref{tab:main}); B-Free scores higher on generator-OOD, for a reason discussed in Section~\ref{sec:res-controls}. None of them has a useful operating point (Table~\ref{tab:controls}). At the max-F1 threshold selected on the validation split, four of the seven flag almost every test image, and all flag between \ZSCtrlFPRMin{} and \ZSCtrlFPRMax{} of the matched controls; at the Youden threshold they flag \ZSYoudenFPRMin{}--\ZSYoudenFPRMax{} of authentic images and controls but detect only \ZSYoudenTPRMin{}--\ZSYoudenTPRMax{} of the manipulations. The control-paired AUROC (Table~\ref{tab:paired}) shows what they do detect. TruFor and Mesorch separate donor-background composites from their resampled controls well (\TruForZSFOnePaired{} and \MesorchZSFOnePaired{} for F1), but not generative backgrounds from autoencoder round trips (\TruForZSFThreePaired{} and \MesorchZSFThreePaired{} for F3): the compositing boundary is a cue they learned from splicing data, a regenerated background is not.

\subsection{Fine-tuning closes most of the gap, with clear differences between detectors}
After fine-tuning on BG-REAL, Mesorch reaches a test-ID AUROC of \FTMesorchID{} and MVSS-Net \FTMVSSID{}; SparseViT stays at \FTSparseID{}. Our reference detector BG-RIFT reaches \FTBGRIFTID{}, on par with Mesorch on test-ID but slightly behind it on the generator- and subject-OOD splits and with a higher control FPR (\FTBGRIFTCtrl{} versus \FTMesorchCtrl{}). The re-implemented background- and subject-forgery detectors remain far behind even with the ground-truth subject mask: BFDNet reaches \BFDNetID{} and SPDNet \SPDNetID{}, with control FPRs of \BFDNetCtrl{} and \SPDNetCtrl{}. Image-level features of the subject and background regions, which suffice for person-centric compositing, do not capture generative and relighting-based edits. Generator-OOD is the hardest held-out condition for every strong fine-tuned detector (MVSS-Net, Mesorch and BG-RIFT, with and without matched-control terms), which lose \DropGenMin{}--\DropGenMax{} AUROC points on it: the held-out generators leave autoencoder traces that differ from the SD1.5 autoencoder seen in training. Held-out scene categories cost the same detectors at most \DropBGMax{} points, held-out subject classes at most \DropSubjMax{} and post-processing (robustness split) at most \DropRobustMax{}. Weaker detectors depend more on the subject class: SparseViT and BFDNet lose 7.5 and 5.6 points on subject-OOD.

\begin{table*}[t]
\centering
\caption{Control-paired AUROC: manipulated images against their own matched controls on the same recipients (F1/C1, F2/C3, F3--F5/C2 on test-ID; G1--G3 against the corresponding autoencoder control on generator-OOD). 0.5 means the detector cannot tell the edit from the processing that accompanies it. $^\ddagger$Detector of AI-generated images.}
\label{tab:paired}
\small
\setlength{\tabcolsep}{4pt}
\resizebox{\linewidth}{!}{%
\begin{tabular}{@{}lcccccccc@{}}
\toprule
Method & F1 & F2 & F3 & F4 & F5 & G1 & G2 & G3 \\
\midrule
\multicolumn{9}{@{}l}{\textit{zero-shot}} \\
TruFor & 0.914 & 0.886 & 0.587 & 0.483 & 0.598 & 0.580 & 0.729 & 0.623 \\
HiFi-Net & 0.463 & 0.465 & 0.495 & 0.593 & 0.465 & 0.500 & 0.546 & 0.487 \\
MVSS-Net & 0.634 & 0.609 & 0.523 & 0.800 & 0.525 & 0.635 & 0.652 & 0.484 \\
IML-ViT & 0.781 & 0.716 & 0.534 & 0.725 & 0.467 & 0.581 & 0.779 & 0.584 \\
Mesorch & 0.892 & 0.853 & 0.549 & 0.781 & 0.663 & 0.566 & 0.758 & 0.550 \\
SparseViT & 0.800 & 0.757 & 0.585 & 0.709 & 0.603 & 0.529 & 0.754 & 0.575 \\
B-Free$^\ddagger$ & 0.804 & 0.779 & 0.794 & 0.871 & 0.113 & 0.880 & 0.772 & 0.876 \\
\midrule
\multicolumn{9}{@{}l}{\textit{fine-tuned}} \\
MVSS-Net & 0.964\,{\scriptsize$\pm$0.012} & 0.953\,{\scriptsize$\pm$0.012} & 0.952\,{\scriptsize$\pm$0.005} & 0.986\,{\scriptsize$\pm$0.008} & 0.704\,{\scriptsize$\pm$0.015} & 0.809\,{\scriptsize$\pm$0.017} & 0.975\,{\scriptsize$\pm$0.010} & 0.818\,{\scriptsize$\pm$0.005} \\
SparseViT & 0.895\,{\scriptsize$\pm$0.006} & 0.884\,{\scriptsize$\pm$0.012} & 0.869\,{\scriptsize$\pm$0.013} & 0.912\,{\scriptsize$\pm$0.022} & 0.611\,{\scriptsize$\pm$0.032} & 0.779\,{\scriptsize$\pm$0.010} & 0.910\,{\scriptsize$\pm$0.022} & 0.751\,{\scriptsize$\pm$0.012} \\
Mesorch & 0.986\,{\scriptsize$\pm$0.002} & 0.983\,{\scriptsize$\pm$0.002} & 0.973\,{\scriptsize$\pm$0.007} & 0.986\,{\scriptsize$\pm$0.010} & 0.777\,{\scriptsize$\pm$0.003} & 0.869\,{\scriptsize$\pm$0.026} & 0.993\,{\scriptsize$\pm$0.005} & 0.840\,{\scriptsize$\pm$0.066} \\
Mesorch + MC & 0.987\,{\scriptsize$\pm$0.002} & 0.984\,{\scriptsize$\pm$0.004} & 0.989\,{\scriptsize$\pm$0.004} & 0.995\,{\scriptsize$\pm$0.001} & 0.860\,{\scriptsize$\pm$0.021} & 0.833\,{\scriptsize$\pm$0.018} & 0.995\,{\scriptsize$\pm$0.002} & 0.869\,{\scriptsize$\pm$0.016} \\
BFDNet$^\dagger$ & 0.802\,{\scriptsize$\pm$0.021} & 0.802\,{\scriptsize$\pm$0.021} & 0.851\,{\scriptsize$\pm$0.027} & 0.890\,{\scriptsize$\pm$0.021} & 0.649\,{\scriptsize$\pm$0.040} & 0.734\,{\scriptsize$\pm$0.034} & 0.921\,{\scriptsize$\pm$0.034} & 0.741\,{\scriptsize$\pm$0.033} \\
SPDNet$^\dagger$ & 0.608\,{\scriptsize$\pm$0.038} & 0.607\,{\scriptsize$\pm$0.037} & 0.789\,{\scriptsize$\pm$0.035} & 0.812\,{\scriptsize$\pm$0.014} & 0.564\,{\scriptsize$\pm$0.045} & 0.647\,{\scriptsize$\pm$0.045} & 0.746\,{\scriptsize$\pm$0.027} & 0.695\,{\scriptsize$\pm$0.044} \\
BG-RIFT & 0.974\,{\scriptsize$\pm$0.004} & 0.969\,{\scriptsize$\pm$0.007} & 0.986\,{\scriptsize$\pm$0.001} & 0.996\,{\scriptsize$\pm$0.000} & 0.806\,{\scriptsize$\pm$0.017} & 0.833\,{\scriptsize$\pm$0.010} & 0.996\,{\scriptsize$\pm$0.003} & 0.875\,{\scriptsize$\pm$0.011} \\
BG-RIFT + MC & 0.977\,{\scriptsize$\pm$0.004} & 0.971\,{\scriptsize$\pm$0.006} & 0.984\,{\scriptsize$\pm$0.004} & 0.997\,{\scriptsize$\pm$0.002} & 0.835\,{\scriptsize$\pm$0.013} & 0.821\,{\scriptsize$\pm$0.011} & 0.995\,{\scriptsize$\pm$0.006} & 0.863\,{\scriptsize$\pm$0.027} \\
 
\bottomrule
\end{tabular}}
\end{table*}

\subsection{What the detectors respond to}
\label{sec:res-controls}
Matched controls make it possible to ask whether a detector responds to the semantic edit or to the processing. Table~\ref{tab:controls} shows that zero-shot detectors flag the controls at the same rate as authentic images, so their false alarms are not caused by a particular processing step but by poor separation of manipulated and authentic images overall. For the strongest fine-tuned detectors (Mesorch, BG-RIFT, MVSS-Net) the control-paired AUROCs of the full-background families F1--F4 exceed 0.95 (Table~\ref{tab:paired}), so they respond to the replaced background and not only to resampling, autoencoding or harmonization; for SparseViT, BFDNet and SPDNet the margin between manipulations and controls is much smaller. This holds for the training generators: against their own SDXL autoencoder controls, SDXL edits (G1) reach paired AUROCs of only \StrongGOnePairedMin{}--\StrongGOnePairedMax{} for these detectors (with and without matched-control terms) and FLUX.1 Fill edits (G3) \StrongGThreePairedMin{}--\StrongGThreePairedMax{}, whereas Kandinsky edits (G2) reach \StrongGTwoPairedMin{}--\StrongGTwoPairedMax{}. The newest generator is thus as hard as SDXL, and a held-out generator is not hard merely because it is recent: Kandinsky's autoencoder trace is apparently easy to tell apart from its edits, SDXL's and FLUX's are not. The local edits (F5) are the exception: they are separated from autoencoder controls with an AUROC of only \FTMesorchFFivePaired{} for fine-tuned Mesorch, because a removed or inserted object changes a small region while the autoencoder trace of the control covers the background.

The AI-generated-image detector B-Free shows why such controls are needed. It is the best zero-shot detector on test-ID (\BFreeID{}) and separates the generator-OOD edits from authentic images almost perfectly (AUROC \BFreeGAuthMin{}--\BFreeGAuthMax{} for G1--G3), which is why its generator-OOD AUROC is the highest of the zero-shot detectors. But at its validation threshold it flags \BFreeFPRCTwo{} of the autoencoder controls of test-ID recipients, against \BFreeFPRAZero{} of authentic images and \BFreeFPRCOne{}--\BFreeFPRCThree{} of the other controls, and against their own autoencoder controls the generator-OOD edits reach only \BFreeGPairedMin{}--\BFreeGPairedMax{}. Local edits are separated from their controls with an AUROC of \BFreeFFivePaired{}: an image with one inserted or removed object looks less generated to B-Free than the same unedited image passed through the autoencoder. B-Free therefore responds to the trace of the diffusion autoencoder, as it was designed to for fully generated images, not to the semantic edit; without processing-matched controls its scores would be read as detection of the edits.

\begin{table*}[t]
\centering
\caption{Operating points on test-ID. Left: true-positive rate on manipulated images and false-positive rate on authentic images (A0) and on each control type at the max-F1 threshold fixed on the validation split. Right: true- and false-positive rate (authentic images and controls) at the threshold maximising Youden's $J$ on the validation split. $^\ddagger$Detector of AI-generated images.}
\label{tab:controls}
\small
\setlength{\tabcolsep}{4pt}
\resizebox{\linewidth}{!}{%
\begin{tabular}{@{}lccccccc@{}}
\toprule
 & \multicolumn{5}{c}{Max-F1 threshold} & \multicolumn{2}{c}{Youden threshold} \\
\cmidrule(lr){2-6}\cmidrule(lr){7-8}
Method & TPR & FPR A0 & FPR C1 & FPR C2 & FPR C3 & TPR & FPR \\
\midrule
TruFor (zero-shot) & 0.935 & 0.819 & 0.769 & 0.854 & 0.849 & 0.598 & 0.289 \\
HiFi-Net (zero-shot) & 0.993 & 0.995 & 0.995 & 0.990 & 0.990 & 0.207 & 0.196 \\
MVSS-Net (zero-shot) & 0.998 & 0.995 & 0.995 & 0.990 & 1.000 & 0.373 & 0.187 \\
IML-ViT (zero-shot) & 1.000 & 1.000 & 1.000 & 1.000 & 1.000 & 0.515 & 0.289 \\
Mesorch (zero-shot) & 0.807 & 0.558 & 0.543 & 0.528 & 0.593 & 0.535 & 0.171 \\
SparseViT (zero-shot) & 0.999 & 0.995 & 0.995 & 1.000 & 0.995 & 0.345 & 0.073 \\
B-Free$^\ddagger$ (zero-shot) & 0.821 & 0.352 & 0.317 & 0.940 & 0.437 & 0.668 & 0.279 \\
MVSS-Net (fine-tuned) & 0.838\,{\scriptsize$\pm$0.007} & 0.119\,{\scriptsize$\pm$0.025} & 0.109\,{\scriptsize$\pm$0.021} & 0.141\,{\scriptsize$\pm$0.023} & 0.124\,{\scriptsize$\pm$0.016} & 0.791\,{\scriptsize$\pm$0.027} & 0.086\,{\scriptsize$\pm$0.034} \\
SparseViT (fine-tuned) & 0.870\,{\scriptsize$\pm$0.044} & 0.303\,{\scriptsize$\pm$0.013} & 0.290\,{\scriptsize$\pm$0.006} & 0.317\,{\scriptsize$\pm$0.027} & 0.317\,{\scriptsize$\pm$0.013} & 0.755\,{\scriptsize$\pm$0.061} & 0.175\,{\scriptsize$\pm$0.033} \\
Mesorch (fine-tuned) & 0.879\,{\scriptsize$\pm$0.034} & 0.055\,{\scriptsize$\pm$0.018} & 0.054\,{\scriptsize$\pm$0.019} & 0.072\,{\scriptsize$\pm$0.029} & 0.054\,{\scriptsize$\pm$0.023} & 0.873\,{\scriptsize$\pm$0.027} & 0.053\,{\scriptsize$\pm$0.013} \\
Mesorch + MC (fine-tuned) & 0.894\,{\scriptsize$\pm$0.004} & 0.067\,{\scriptsize$\pm$0.003} & 0.059\,{\scriptsize$\pm$0.003} & 0.047\,{\scriptsize$\pm$0.018} & 0.070\,{\scriptsize$\pm$0.005} & 0.890\,{\scriptsize$\pm$0.007} & 0.056\,{\scriptsize$\pm$0.009} \\
BFDNet$^\dagger$ (fine-tuned) & 0.845\,{\scriptsize$\pm$0.012} & 0.379\,{\scriptsize$\pm$0.035} & 0.380\,{\scriptsize$\pm$0.038} & 0.369\,{\scriptsize$\pm$0.029} & 0.382\,{\scriptsize$\pm$0.035} & 0.755\,{\scriptsize$\pm$0.052} & 0.261\,{\scriptsize$\pm$0.038} \\
SPDNet$^\dagger$ (fine-tuned) & 0.981\,{\scriptsize$\pm$0.023} & 0.946\,{\scriptsize$\pm$0.076} & 0.948\,{\scriptsize$\pm$0.069} & 0.940\,{\scriptsize$\pm$0.083} & 0.951\,{\scriptsize$\pm$0.067} & 0.542\,{\scriptsize$\pm$0.072} & 0.272\,{\scriptsize$\pm$0.080} \\
BG-RIFT (fine-tuned) & 0.880\,{\scriptsize$\pm$0.009} & 0.097\,{\scriptsize$\pm$0.008} & 0.090\,{\scriptsize$\pm$0.010} & 0.109\,{\scriptsize$\pm$0.023} & 0.097\,{\scriptsize$\pm$0.008} & 0.863\,{\scriptsize$\pm$0.008} & 0.075\,{\scriptsize$\pm$0.014} \\
BG-RIFT + MC (fine-tuned) & 0.879\,{\scriptsize$\pm$0.019} & 0.080\,{\scriptsize$\pm$0.018} & 0.079\,{\scriptsize$\pm$0.016} & 0.085\,{\scriptsize$\pm$0.018} & 0.080\,{\scriptsize$\pm$0.017} & 0.851\,{\scriptsize$\pm$0.019} & 0.059\,{\scriptsize$\pm$0.024} \\
 
\bottomrule
\end{tabular}}
\end{table*}

\subsection{Localisation: inverted masks and why per-image F1 misleads}
\begin{figure*}[t]
\centering
\includegraphics[width=\linewidth]{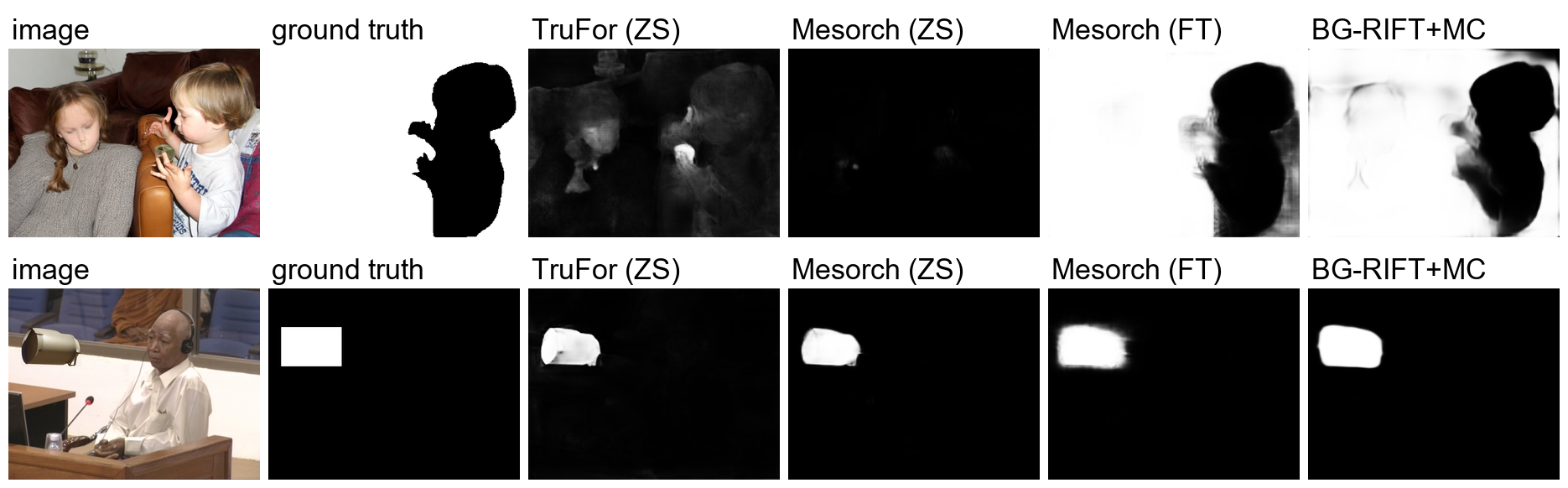}
\caption{Localisation on test-ID. Top: generative background replacement (F3); the edit mask is the background. Bottom: object insertion into the background (F5). Zero-shot detectors highlight the subject or the inserted object, never a replaced background; fine-tuned detectors segment the replaced background. The manipulated images were produced by the generators listed in Table~\ref{tab:families} as part of the benchmark.}
\label{fig:loc}
\end{figure*}

Zero-shot localisation is not merely weak but inverted: Matthews correlations on test-ID are negative for every zero-shot localiser (B-Free gives no mask), down to \ZSMCCMin{} (Table~\ref{tab:loc}). Trained on splicing, the detectors mark the region that looks pasted---here the preserved subject---as manipulated (Fig.~\ref{fig:loc}, top). Inserted objects, which do resemble spliced objects, are found (Fig.~\ref{fig:loc}, bottom). Fine-tuned detectors segment the replaced background (MCC between 0.74 and 0.82). The segmentation diagnostic puts these numbers in perspective: predicting the complement of the ground-truth subject gives a per-image pixel F1 of \SegPixFOne{} on manipulated test-ID images and exactly 1 on every full-background family, because the edit mask of a full replacement \emph{is} the complement of the subject. Once false alarms on authentic and control images are counted, its pooled pixel F1 drops to \SegPixFOneAll{}, and on local edits, whose location is not given by the subject, it reaches only \SegFFivePixF{}. Localisation on full-background replacements therefore mostly measures subject segmentation; the informative localisation numbers are the pooled F1, which rewards silence on authentic images, and pixel F1 on F5, where the best detector reaches \FTMesorchFFivePixF{}.

\begin{table}[t]
\centering
\caption{Localisation on test-ID: pixel F1 on local edits (F5), Matthews correlation and tolerant boundary F1 on manipulated images, and pixel F1 pooled over all images (false alarms on authentic images and controls count).}
\label{tab:loc}
\small
\setlength{\tabcolsep}{3pt}
\resizebox{\linewidth}{!}{%
\begin{tabular}{@{}lcccc@{}}
\toprule
Method & F5 F1 & MCC & Bnd.\ F1 & Pooled F1 \\
\midrule
TruFor (zero-shot) & 0.273 & -0.396 & 0.400 & 0.038 \\
HiFi-Net (zero-shot) & 0.001 & -0.003 & 0.002 & 0.071 \\
MVSS-Net (zero-shot) & 0.039 & -0.191 & 0.124 & 0.023 \\
IML-ViT (zero-shot) & 0.238 & -0.421 & 0.429 & 0.051 \\
Mesorch (zero-shot) & 0.234 & -0.379 & 0.383 & 0.014 \\
SparseViT (zero-shot) & 0.150 & -0.268 & 0.279 & 0.015 \\
MVSS-Net (fine-tuned) & 0.267\,{\scriptsize$\pm$0.012} & 0.746\,{\scriptsize$\pm$0.003} & 0.576\,{\scriptsize$\pm$0.003} & 0.827\,{\scriptsize$\pm$0.006} \\
SparseViT (fine-tuned) & 0.165\,{\scriptsize$\pm$0.109} & 0.738\,{\scriptsize$\pm$0.054} & 0.623\,{\scriptsize$\pm$0.058} & 0.812\,{\scriptsize$\pm$0.017} \\
Mesorch (fine-tuned) & 0.436\,{\scriptsize$\pm$0.022} & 0.818\,{\scriptsize$\pm$0.059} & 0.704\,{\scriptsize$\pm$0.064} & 0.940\,{\scriptsize$\pm$0.021} \\
Mesorch + MC (fine-tuned) & 0.335\,{\scriptsize$\pm$0.084} & 0.811\,{\scriptsize$\pm$0.018} & 0.677\,{\scriptsize$\pm$0.015} & 0.953\,{\scriptsize$\pm$0.006} \\
BG-RIFT (fine-tuned) & 0.426\,{\scriptsize$\pm$0.014} & 0.794\,{\scriptsize$\pm$0.009} & 0.665\,{\scriptsize$\pm$0.011} & 0.913\,{\scriptsize$\pm$0.004} \\
BG-RIFT + MC (fine-tuned) & 0.429\,{\scriptsize$\pm$0.007} & 0.779\,{\scriptsize$\pm$0.012} & 0.645\,{\scriptsize$\pm$0.013} & 0.914\,{\scriptsize$\pm$0.005} \\
\midrule
Segmentation only (oracle subject) & 0.080 & 0.825 & 0.806 & 0.621 \\
 
\bottomrule
\end{tabular}}
\end{table}

\subsection{File-level shortcuts}
A logistic regression on file-level properties alone reaches a test-ID AUROC of \MetaAUROCvTwo{} on BG-REAL. Because every recipient's samples share format and JPEG quality, this residual signal comes from image content---regenerated and composited backgrounds compress to a different number of bits at the same quality---and it does not distinguish controls from authentic images: at a threshold of 0.5 the classifier flags \MetaFlagAuthTwo{} of authentic test-ID images and \MetaFlagCtrlTwo{} of their controls. On the first release of this benchmark, where authentic images were stored losslessly and most manipulations and all controls as JPEG, the same diagnostic reaches an AUROC of \MetaAUROCvOne{} and flags \MetaFlagAuthOne{} of authentic images but \MetaFlagCtrlOne{} of controls. A large part of what an earlier matched-control analysis attributed to re-encoding sensitivity was therefore a file-format difference, which the per-recipient encoding of BG-REAL removes by design.

\subsection{Matched-control training}
\label{sec:res-mc}
Table~\ref{tab:mc} compares each detector trained with and without the matched-control terms; everything else, including the triplet batches, is identical. Test-ID AUROC and the control false-positive rate are unchanged for both detectors. Of the twelve comparisons in the table, two remain significant after Holm correction over all twelve \cite{holm1979}, one per detector: for BG-RIFT the terms raise the subject-OOD AUROC by \MCBGRIFTSubjAUROC{} (95\% CI \MCBGRIFTSubjAUROCCI, $p_{\text{Holm}}=\MCBGRIFTSubjAUROCPHolm$), and for Mesorch they raise the control-paired AUROC of local edits from \FTMesorchFFivePaired{} to \FTMesorchMCFFivePaired{} (\MCMesorchFFive{} \MCMesorchFFiveCI, $p_{\text{Holm}}=\MCMesorchFFivePHolm$). One further effect excludes zero only without correction: BG-RIFT's false-positive rate on subject-OOD authentic images and controls falls by \MCBGRIFTSubjFPR{} \MCBGRIFTSubjFPRCI{} ($p_{\text{Holm}}=\MCBGRIFTSubjFPRPHolm$). On generator-OOD the terms have no net effect for either detector (Mesorch \MCMesorchGenAUROC{} \MCMesorchGenAUROCCI{}, BG-RIFT \MCBGRIFTGenAUROC{} \MCBGRIFTGenAUROCCI{}), but the per-generator columns of Table~\ref{tab:paired} show opposite effects that cancel: for Mesorch the terms lower the paired AUROC on SDXL edits and raise it on FLUX.1 Fill edits. A plausible reading is that asking the detector to treat the SD1.5 autoencoder trace---shared by the controls and most training manipulations---as uninformative makes it rely more on the edit itself, which helps where that trace is uninformative (local edits, unfamiliar subjects) and can hurt where it partly transfers to another autoencoder. With two detectors and three seeds each, we present the terms as a cheap, model-agnostic option with detector-dependent benefits, not as a uniform improvement.

\begin{table}[t]
\centering
\caption{Effect of the matched-control terms (with $-$ without), pooled over three paired seeds, with 95\% bootstrap confidence intervals over test recipients. Bold: interval excludes zero; $^{*}$: significant after Holm correction over all twelve comparisons (two-sided bootstrap $p$-values). For FPR, negative is better.}
\label{tab:mc}
\small
\setlength{\tabcolsep}{3pt}
\begin{tabular}{@{}lcc@{}}
\toprule
Metric & BG-RIFT & Mesorch \\
\midrule
AUROC ID & +0.004 [-0.002, +0.013] & +0.007 [-0.007, +0.019] \\
AUROC Gen-OOD & -0.009 [-0.032, +0.010] & -0.004 [-0.032, +0.037] \\
AUROC Subj-OOD & \textbf{+0.023 [+0.007, +0.040]}$^{*}$ & +0.014 [-0.002, +0.040] \\
Paired AUROC F5 & +0.028 [-0.007, +0.069] & \textbf{+0.083 [+0.035, +0.135]}$^{*}$ \\
FPR controls & -0.017 [-0.053, +0.022] & -0.001 [-0.053, +0.040] \\
FPR Subj-OOD & \textbf{-0.118 [-0.245, -0.012]} & +0.001 [-0.102, +0.071] \\
 
\bottomrule
\end{tabular}
\end{table}

\subsection{Ablation of BG-RIFT}
Table~\ref{tab:ablation} removes one component of BG-RIFT at a time. ImageNet pretraining is essential: without it the test-ID AUROC falls to \NoPretrainID{}. The SRM residual stream and the region supervision, in contrast, do not help measurably (without the residual stream the AUROC is \NoResID{}); within the variance of three seeds the RGB encoder alone carries the detection. We report BG-RIFT as a compact reference model that reaches the accuracy of the strongest fine-tuned detector on test-ID, not as an architectural contribution.

\begin{table}[t]
\centering
\caption{Ablation of BG-RIFT trained with the matched-control terms (mean $\pm$ std over three seeds).}
\label{tab:ablation}
\small
\setlength{\tabcolsep}{3pt}
\begin{tabular}{@{}lcccc@{}}
\toprule
Variant & ID & Gen-OOD & Ctrl FPR & F5 F1 \\
\midrule
BG-RIFT + MC (full) & 0.949\,{\scriptsize$\pm$0.002} & 0.893\,{\scriptsize$\pm$0.011} & 0.082\,{\scriptsize$\pm$0.017} & 0.429\,{\scriptsize$\pm$0.007} \\
-- SRM residual stream & 0.951\,{\scriptsize$\pm$0.004} & 0.901\,{\scriptsize$\pm$0.014} & 0.080\,{\scriptsize$\pm$0.036} & 0.426\,{\scriptsize$\pm$0.015} \\
-- region supervision & 0.946\,{\scriptsize$\pm$0.003} & 0.905\,{\scriptsize$\pm$0.006} & 0.107\,{\scriptsize$\pm$0.040} & 0.417\,{\scriptsize$\pm$0.004} \\
-- ImageNet pretraining & 0.772\,{\scriptsize$\pm$0.006} & 0.768\,{\scriptsize$\pm$0.009} & 0.623\,{\scriptsize$\pm$0.054} & 0.049\,{\scriptsize$\pm$0.003} \\
-- matched-control terms & 0.945\,{\scriptsize$\pm$0.001} & 0.902\,{\scriptsize$\pm$0.002} & 0.099\,{\scriptsize$\pm$0.012} & 0.426\,{\scriptsize$\pm$0.014} \\
 
\bottomrule
\end{tabular}
\end{table}

\section{Limitations}
\label{sec:limitations}
\paragraph{Generators and edits} Training manipulations come from one diffusion family (SD1.5 inpainting and the SD1.5-based IC-Light) plus PCT-Net and LaMa; generator-OOD covers three further generators, one of them a flow-matching transformer (FLUX.1 Fill), which we ran with 4-bit weights to fit the GPU and which may therefore leave slightly different traces than at full precision. Commercial editors, other recent diffusion and flow-matching models, video-conferencing pipelines and manual edits by skilled users are not included, and the generator-OOD results should not be read as generalisation to all unseen generators.

\paragraph{Localisation target} For full-background replacement the semantic edit mask is the complement of the subject, so per-image localisation scores on these families mostly measure subject segmentation (Section~\ref{sec:results}). We report pooled pixel F1 and local-edit localisation for this reason, but the benchmark is primarily a detection benchmark.

\paragraph{Quality control} Quality control is automatic: it removes failed generations, unrestored subjects and edits that did not change their region, but it does not rate visual realism. Some composites are recognisably unnatural (e.g.\ lighting or perspective mismatch in F1, floating inserted objects in F5); such samples are realistic for low-effort manipulations but make parts of the benchmark easier than skilled edits would be. Recipient images may contain secondary people or objects, which become part of the replaced background. No human study was conducted for this release.

\paragraph{Compared methods} BFDNet and SPDNet were re-implemented from their papers and given the ground-truth subject mask; their numbers may differ from the authors' implementations. CAT-Net and PSCC-Net could not be run in our environment, and the BR-Gen data were not accessible at the time of writing, so we could not test cross-dataset transfer between BR-Gen and BG-REAL. Detectors designed for AI-generated local edits, including multimodal-LLM-based localisers \cite{xu2025fakeshield,huang2025sida} and the detector released with BR-Gen \cite{cai2026brgen}, are not evaluated; of the detectors for AI-generated images we evaluate only B-Free, zero-shot. All training used one 8~GB GPU, which limited input resolution to 512 pixels, the training budget to three or four epochs, and the number of seeds to three. MVSS-Net was still improving at the end of fine-tuning (its validation AUROC rose by up to 0.025 in the last epoch), so its results may understate what longer training achieves; the validation AUROC of Mesorch and BG-RIFT had levelled off.

\paragraph{Data and ethics} All source and donor photographs are Open Images V7 images licensed under CC BY 2.0; we redistribute them and their edited versions with per-image attribution (author, title and original URL). Images showing people were not selected for identity, and the benchmark is intended for building and auditing detectors, not for producing manipulations.

\section{Conclusion}
\label{sec:conclusion}
BG-REAL evaluates detectors on subject-preserving background manipulations---donor compositing, learned harmonization, generative inpainting, relighting and local background edits---and pairs every processing path with an authentic control. Three findings stand out. First, detectors released for general manipulation localisation do not transfer to this setting: zero-shot AUROCs stay at or below \ZSAUROCMax{}, none has a useful operating point, and their localisation is inverted, marking the preserved subject instead of the replaced background. A detector of AI-generated images does better on raw AUROC but responds to the diffusion autoencoder, not to the edit: it flags most authentic images that only passed through the autoencoder. Second, fine-tuning on BG-REAL raises detection to an AUROC of about 0.95 for the strongest detectors, and matched controls show that these detectors respond to the edit rather than to the processing for full-background replacements made with the training generators, but much less so for local edits and for SDXL and FLUX.1 Fill edits, the latter from the newest generator in the benchmark. Third, using the controls during training leaves in-distribution accuracy unchanged and has detector-dependent effects: after correction for multiple comparisons it improved detection on unseen subjects for BG-RIFT and on local edits for Mesorch, while its effect on unseen generators differed between generators and cancelled out overall. For practitioners, the controls and the pooled localisation measure provide checks that standard accuracy tables omit; for future benchmarks, the observation that full-background localisation reduces to subject segmentation calls for detection-centred evaluation of such edits. Extending the generator pool, adding skilled human edits and testing transfer across background-editing datasets are the next steps.

\section*{Data availability}
The BG-REAL images, masks, splits, generation metadata (generator repository revisions, prompts, seeds, donor identifiers), per-sample predictions and evaluation tables, and trained weights are deposited at \url{https://huggingface.co/datasets/NoirZangetsu/bg-real} (folder \texttt{v2}; the first release is preserved at tag \texttt{v1-imavis}). Fine-tuned MVSS-Net weights are not redistributed because the upstream repository carries no licence. Source and donor photographs are Open Images V7 images licensed under CC BY 2.0 and are redistributed with per-image attribution. Generators were used under their respective licences (CreativeML OpenRAIL-M, OpenRAIL++-M, Apache-2.0, MPL-2.0, and the FLUX.1 [dev] Non-Commercial License, to whose terms the G3 images are subject); the TruFor and B-Free weights are used under their non-commercial research licences.

\section*{Author contributions}
Bugra Alperen Uluirmak: Conceptualization, Methodology, Software, Data curation, Formal analysis, Investigation, Writing -- original draft. Rifat Kurban: Writing -- review \& editing.

\section*{Declaration of competing interest}
The authors declare no competing financial interests or personal relationships that could have appeared to influence the work reported in this paper.

\section*{Funding}
This research did not receive any specific grant from funding agencies in the public, commercial, or not-for-profit sectors.

\section*{Declaration of generative AI and AI-assisted technologies in the manuscript preparation process}
During the preparation of this work the authors used Claude (Anthropic) to correct and edit the manuscript text. After using this tool, the authors reviewed and edited the content as needed and take full responsibility for the content of the published article.

\bibliographystyle{elsarticle-num}
\bibliography{refs}

\end{document}